\pdfoutput = 1
\documentclass[letterpaper, 10pt, conference]{ieeeconf}

\IEEEoverridecommandlockouts  
\makeatletter
\let\NAT@parse\undefined
\makeatother

\usepackage[pdftex]{graphics} 
\usepackage{epsfig} 
\usepackage[cmex10]{amsmath} 
\usepackage{newtxmath} 
\usepackage{times} 
\usepackage[hidelinks]{hyperref} 
\usepackage[all]{hypcap} 
\usepackage[capitalise,nameinlink]{cleveref} 
\usepackage{algorithm}
\usepackage{algpseudocode}
\usepackage{todonotes}
\usepackage{bm}
\usepackage{multirow}
\usepackage{subcaption}
\usepackage{caption}
\usepackage{array}
\usepackage{pifont}
\newcommand{\cmark}{\ding{51}}%
\newcommand{\xmark}{\ding{55}}%

\usepackage{booktabs}

\newcommand\norm[1]{\left\lVert#1\right\rVert}

\title{\LARGE \bf Reaching, Grasping and Re-grasping: Learning Multimode Grasping Skills}

\author{Wenbin Hu, Chuanyu Yang, Kai Yuan, Zhibin Li
\thanks{All authors are with the School of Informatics, The University of Edinburgh (10 Crichton Street, Edinburgh, EH8 9AB, United Kingdom)
\newline Email: {\tt\small wenbin.hu@ed.ac.uk}}
}

\graphicspath{{./images/}}

\begin{document}

\bstctlcite{IEEEexample:BSTcontrol}
\bibliographystyle{IEEEtran}
\maketitle

\begin{abstract}
The ability to adapt to uncertainties, recover from failures, and coordinate between hand and fingers are essential sensorimotor skills for fully autonomous robotic grasping. 
In this paper, we aim to study a unified feedback control policy for generating the finger actions and the motion of hand to accomplish seamlessly coordinated tasks of reaching, grasping and re-grasping. We proposed a set of quantified metrics for task-orientated rewards to guide the policy exploration, and we analyzed and demonstrated the effectiveness of each reward term. To acquire a robust re-grasping motion, we deployed different initial states in training to experience failures that the robot would encounter during grasping due to inaccurate perception or disturbances. The performance of learned policy is evaluated on three different tasks: grasping a static target, grasping a dynamic target, and re-grasping. The quality of learned grasping policy was evaluated based on success rates in different scenarios and the recovery time from failures. The results indicate that the learned policy is able to achieve stable grasps of a static or moving object. Moreover, the policy can adapt to new environmental changes on the fly and execute collision-free re-grasp after a failed attempt within a short recovery time even in difficult configurations.
\end{abstract}

\section{Introduction}\label{seq:introduction}
Reactive adaption to new changes and recovery from failures are important features for any control policy for real-world robot applications in the future. For this, autonomous grasping is the fundamental capability of many robotic manipulation tasks. However, the combined control of reaching, grasping and re-grasping in a dynamically changing, non-stationary environment remains a challenge.

In the traditional planning approaches, reaching and grasping are inherently different and usually planned separately and deployed sequentially. For grasping of a moving ball, vision and proximity sensors have been used from a top-view \cite{2015_Suzuki}. Marturi et al. developed an approach of planning pre-grasp posture online and tracking a moving object, where the grasp motion was determined by a human operator \cite{2019_Marturi}. Planning of the complete reaching and grasping motion is quite time-consuming and is often implemented in an open-loop or partially reactive controlled manner \cite{2010_KROEMER}. Generally, current planning based methods have good results in solving reaching \cite{2015_Suzuki} or grasping problem \cite{2017_Hang} individually, but the switch between controllers is designed manually. As a next-level performance with increased robustness, reaching, grasping and even re-grasping \textit{should be addressed simultaneously in one unified policy}.

Machine learning methods provide a promising option for autonomous control, as they alleviate the requirements of manual design and prior knowledge as they can autonomously explore the whole operation space and react in potential corner cases. Recently, vision-based data-driven methods show prominent performances \cite{2014_Bohg, 2018_Lu} in the optimization of grasping static objects. Compared with classical grasp synthesis, learning-based approaches improve the performance of grasping unknown objects dramatically \cite{2015_Kappler,2015_Pinto,2016_Levine, 2017_Mahler}. However, training of the model requires very large data, either collected from simulation \cite{2015_Kappler} or self-supervised real robot experiment which is time-consuming \cite{2015_Pinto}. Furthermore, sampling and ranking of grasp candidates often takes long computation time \cite{2015_Pinto, 2017_Mahler}, which limits the capability of reactive control. The success of a grasp strongly relies on precise object perception and accurate hardware control. In case of a failure, no recovery strategy is being deployed, and the whole pipeline is reset and another attempt is repeated instead of an on-line, reactive adjustment \cite{2015_Kappler}.


\begin{figure}[t]
	\centering
	\captionsetup{width=1.0\linewidth}
	\includegraphics[width=60mm]{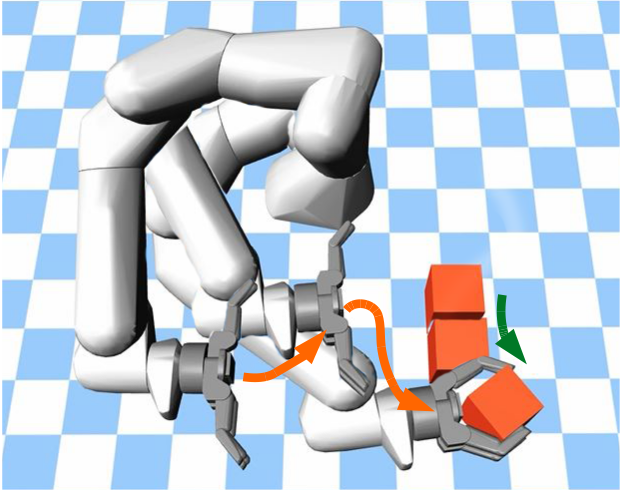}
	\caption{Grasping a moving object.} 
	\label{fig:moving_grasp}
	\vspace{-8mm}
\end{figure}

Recent research in Deep Reinforcement Learning (DRL) has shown promising capabilities of solving continuous control tasks with high-dimensional state and action spaces, such as pouring liquids \cite{2015_Hangl}, multi-finger grasping \cite{2019_ICRA_mbkrb}, in-hand manipulation \cite{2018_OpenAI} or bipedal locomotion tasks \cite{2018_Yang}.
In a Reinforcement Learning framework, an agent learns a policy from scratch by maximizing the expected cumulative return from autonomous interactions with environment. In contrast to other Machine Learning techniques, such as unsupervised and supervised learning, no pre-collected training data is required as the agent autonomously generates the training data by interacting with the environment, and infers the quality of its state and actions through reward signals. 
Not requiring pre-generated training data is especially useful in large continuous action and state spaces, because labelling whether one action under current state is good or bad is infeasible due to the infinite amount of possible combinations. 


The focus of this paper is to study a unified control policy for reaching, grasping and re-grasping, which requires synergetic behaviours, fine coordination between hand and fingers. This unified control policy is obtained from training an agent through DRL without any human interference or hard-coded control architecture. 
Although the task spaces for hand and fingers are different, they have to observe each other's state and coordinate properly, especially when the hand is approaching the object, otherwise unwanted collision might happen. 
In this work, we are neither aiming to benchmark with the reaching ability of planning methods nor with the grasp quality of cutting-edge data-driven methods. Instead, we intend to \textbf{learn a unified policy with coordinated motor skills for the entire grasping loop}. Most importantly, the policy should be capable of \textbf{re-grasping quickly in case of failure}; a problem which has been partially-addressed in aforementioned methods. Real-time failure recovery and re-grasp can significantly increase the robustness and efficiency, and further enable real-world applications, compared to the methods which simply restarts the whole pipeline.

The main contributions of this paper are threefold:
\begin{itemize}
\item[(1)] A unified policy of reaching, grasping and re-grasping learned from Deep Reinforcement Learning.
\item[(2)] A task-orientated reward function and special initial states for learning a robust policy.
\item[(3)] Learned policy that is able to achieve robust grasp of static or moving objects, adjust its motion on-line under sudden changes, and re-grasp quickly after failures, even in some challenging configurations, as shown in \cref{fig:moving_grasp}.
\end{itemize}

In the remainder of this paper, we first discuss the related work in the next section. We elaborate the reinforcement learning algorithm in \cref{seq:preliminary}. The details of the simulation design for policy learning are presented in \cref{seq:policy_learning}. The results of learned policies are analyzed and evaluated in \cref{seq:evaluation}. Finally we summarize the paper and suggest future work in \cref{seq:conclusion}.

\section{Related Work}\label{seq:related_work}

Visual servoing methods \cite{1993_vs} offer a proper solution to the integration of reaching and grasping. The real-time action is determined by the current vision input, so that the visual servoing methods inherently have the capability of reactive control. However, most applications require large amounts of prior knowledge about the environment and the task \cite{1997_won}, or complex hierarchical control architectures \cite{2003_akio}.

According to the grasping task, force \cite{2011_Pastor} or tactile information \cite{2018_Roberto,2018_Francois,2016_Chebotar} has been used as feedback to close the control loop and guide the re-grasping motion. These approaches, however, only modify the applied finger forces or the hand posture for a static target, but are not addressing the problem of handling moving object, or the coordination between hand and fingers.

Some researches solve the reaching and grasping problem through a combination of trajectory planning with policy learning \cite{2010_KROEMER} or imitation learning \cite{2007_Ratliff}. These methods achieved good results on grasping static novel targets, but the requirements of prior knowledge and hand-crafted control architecture limit the capability of handling environmental changes, such as the sudden movement of the object.

In order to reduce the reliable of knowledge about the system and task's solution, Lampe et al. \cite{2013_Lampe} combine the classic visual servoing method with DRL. The controller is learned from scratch by success or failure, to control the robot arm to reach and grasp a moving bowl on the table from top-down. The entire combined system is split into two parts: long-range controller mainly for reaching and short-range controller for more precise motion of reaching and grasping. The two controllers are generated differently and use different cameras as vision input. Also, the switch of them is triggered by hand-crafted condition. Compared to \cite{2013_Lampe}, instead of manual partition of controller, our approach learns the policy of reaching and grasping in a holistic manner as one policy.

One major deficiency of learning-based grasp detection methods is the long computation time caused by large CNN and individually sampled and ranked grasp candidates \cite{2015_Pinto, 2017_Mahler}. Morrison et al. overcome the problem by proposing a lightweight network structure that enables reactive close-loop control \cite{2018_Morrison}. The learned controller can dynamically track and grasp novel objects in clutter and achieve high success rate. However, this approach has not yet considered the re-grasp problem in case of a failed attempt. In our approach, we introduce challenging configurations that cause failure grasps as additional initial states to train the collision-free re-grasp motion.

\section{Preliminaries} \label{seq:preliminary}
In this section we briefly introduce deep reinforcement learning and the proximal policy optimization algorithm that we use for problem formulation and policy learning.
\subsection{Reinforcement Learning(RL)}
The task of reaching and grasping an object is considered as a finite-horizon discounted Markov decision process (MDP), consisting of a state space $\mathcal{S}$, an action space $\mathcal{A}$, a distribution of initial states $p(s_0)$, the state transition dynamics $\mathcal{T}:\mathcal{S}\times\mathcal{A}\rightarrow{}\mathcal{S}$, a reward function $r: \mathcal{S}\times\mathcal{A}\rightarrow{}\mathbb{R}$, and a discount factor $\gamma\in(0,1]$. Every learning episode starts with a sampled initial state $s_0$. Thereafter, at every timestep, the agent chooses one action based on current state and the policy $\pi(s_t)$ to be executed. After execution, the agent will receive a reward $r(s_t,a_t)$ and the state observation $s_{t+1}$ from the environment. The goal of the agent is to maximize the expected discounted sum of rewards $\varmathbb{E}_\pi\left[\sum_{t=0}^{T-1}\gamma ^tr(s_t,a_t)\right]$.

\subsection{Proximal Policy Optimization(PPO)}
In this work, we use an on-policy deep reinforcement learning algorithm named Proximal Policy Optimization (PPO) \cite{2017_Schulman} for policy optimization. We implement PPO in an actor-critic fashion, with the actor consisting of a policy $\pi_{\theta}(s_t)$ parameterized by $\theta$ and a critic consisting of estimated value function $V_{\phi}(s_t)$ parameterized by $\phi$.
			
The objective function of PPO is
\begin{equation}\label{eq:objective_function}
    L(\theta)=\hat{\varmathbb{E}_t}\left[\text{min}\left(r_t(\theta)\hat{A_t},\text{clip}(r_t(\theta),1-\epsilon,1+\epsilon)\hat{A_t}\right)\right],
\end{equation}
where $r_t(\theta)$ denotes the probability ratio $\frac{\pi(a_t|s_t)}{\pi_{\text{old}}(a_t|s_t)}$ and $\hat{A_t}$ denotes the estimate of advantage value suggesting whether the action $a_t$ is better or worse than the average action the policy takes at $s_t$. $\epsilon$ is a hyperparameter designed to clip the probability ratio and constrain the policy update. This objective function allows the policy to update towards action distributions with positive advantage while avoiding excessively large policy changes.
			
The goal of PPO is to maximize the objective function $L(\theta)$, therefore $\pi_{\theta}$ is updated by gradient ascent w.r.t \cref{eq:objective_function}. The estimated value function $V_{\phi}$ is trained by minimizing the loss function:
\begin{equation}\label{eq:loss_function}
L(\phi)=\hat{\varmathbb{E}_t}\left[(V(s_t) - R_{t})^{2}\right],
R_{t} = \sum_{l=0}^{T-t}\gamma^{l}r_{t+l},
\end{equation}
where $R_{t}$ is the discounted reward during timestep $t$, $\gamma$ is the discount factor and $T$ is the total number of timestep during the samped path. $V_{\phi}$ is updated by gradient descent w.r.t \cref{eq:loss_function}. Both of the policy $\pi_{\theta}$ and the value function $V_{\phi}$ are parameterized with a fully-connected neural network with two hidden layers of 64 units.

\section{Policy Learning of Dynamic Grasping} \label{seq:policy_learning}
In this section, we present the details of learning a unified policy for reaching, grasping and re-grasping in simulation as demonstrated in \cref{alg:learning}. First, we introduce the simulation environment. Afterwards we describe the control framework. Then, we explain the definition of the state and action space, as well as the design of the reward. Finally, we introduce the structure of training episodes.
\begin{algorithm}[t]
\caption{Policy learning for dynamic grasping.}
\label{alg:learning}
\begin{algorithmic}[1]
\For{$k \in \{1,...,N\}$}                    
    \If{$rand(0,1)<\beta$}
        \State {Normal initial state with random object position}
    \Else
        \State {One special initial state in \cref{fig:initial_states}}
    \EndIf
    \For{$t \in \{1,...,T\}$}
        \State {Get the current state $\mathcal{S}=\{\bm{X_r}, \theta, \bm{q}, \bm{d}, \bm{F}\}$}
        \State {Run policy $\pi_\theta$ and get the action $a_t$}
        \State {Execute $a_t$ based on low-level controller in \cref{fig:control_framework}}
        \State {Compute reward $r_t$ with \cref{eq:reward_function}}
        \State {Collecting tuple $\{s_t, a_t, r_t\}$}
    \EndFor
    \State {$\pi_{\text{old}}\leftarrow\pi_\theta$}
    \State {Update $\theta$ by stochastic gradient ascent w.r.t. \cref{eq:objective_function}}
    \State {Update $\phi$ by stochastic gradient descent w.r.t. \cref{eq:loss_function}}
\EndFor
\end{algorithmic}
\end{algorithm}

\subsection{Simulation Setup} \label{seq:simulation_setup}
\begin{figure}[t!]
\centering
\captionsetup{width=\linewidth}
\includegraphics[trim=300 150 400 150, clip, width=0.6\linewidth]{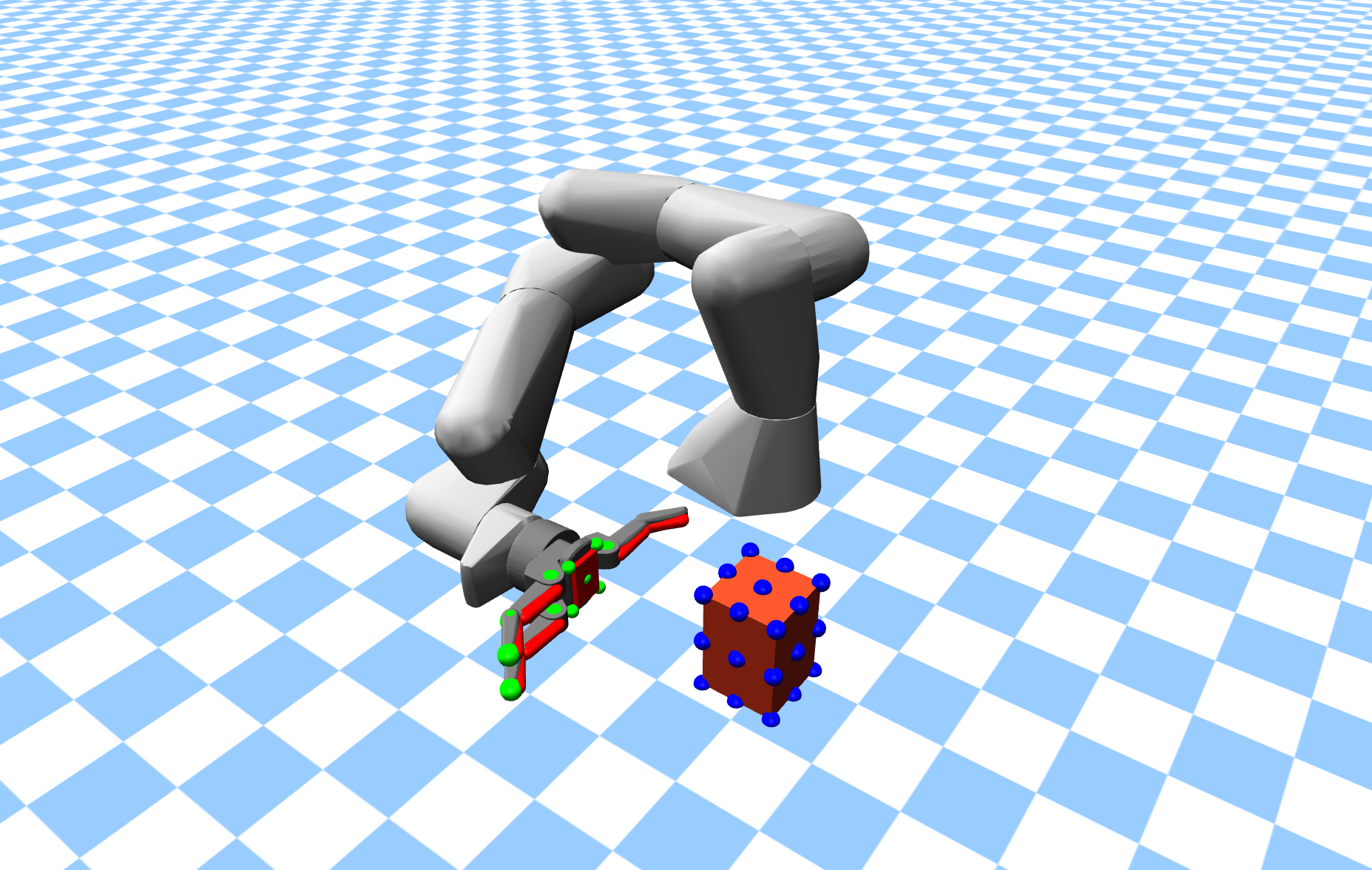}
\caption{Barrett Hand attached on the Franka Emika arm and the target object placed on the ground. Blue points: geometry key points of object's bounding box. Green points: vertexes for forming hand convex hull. Red regions on hand: virtual sites for detecting contact and attaching force sensors. } 
\label{fig:simulation_setup}
\vspace{-2mm}
\end{figure}

For simulating stable, realistic contacts and dynamics, we use the physics simulation engine MuJoCo \cite{2012_Todorov}. We use the Barrett Hand as the end-effector, attached on the Franka Emika robot arm. For policy training, the target object is a cube, and the simulation setup is in \cref{fig:simulation_setup}.

In simulation, the agent will learn the proper motion of the robot hand, including the translational and rotational hand movement and the finger actions to accomplish the combined task of reaching, grasping and re-grasping of an object.
The action space only involves the motion of the end-effector. The motion of the arm is generated through off-the-shelf inverse kinematics solver and motion planner. Since we set early termination signal of self-collision and hitting joint limit, and the object is placed within a limited workspace, the learned motion is safe and collision-free.

\begin{figure}[t]
	\centering
	\captionsetup{width=\linewidth}
	\includegraphics[trim=150 400 250 300,clip,width=0.8\linewidth]{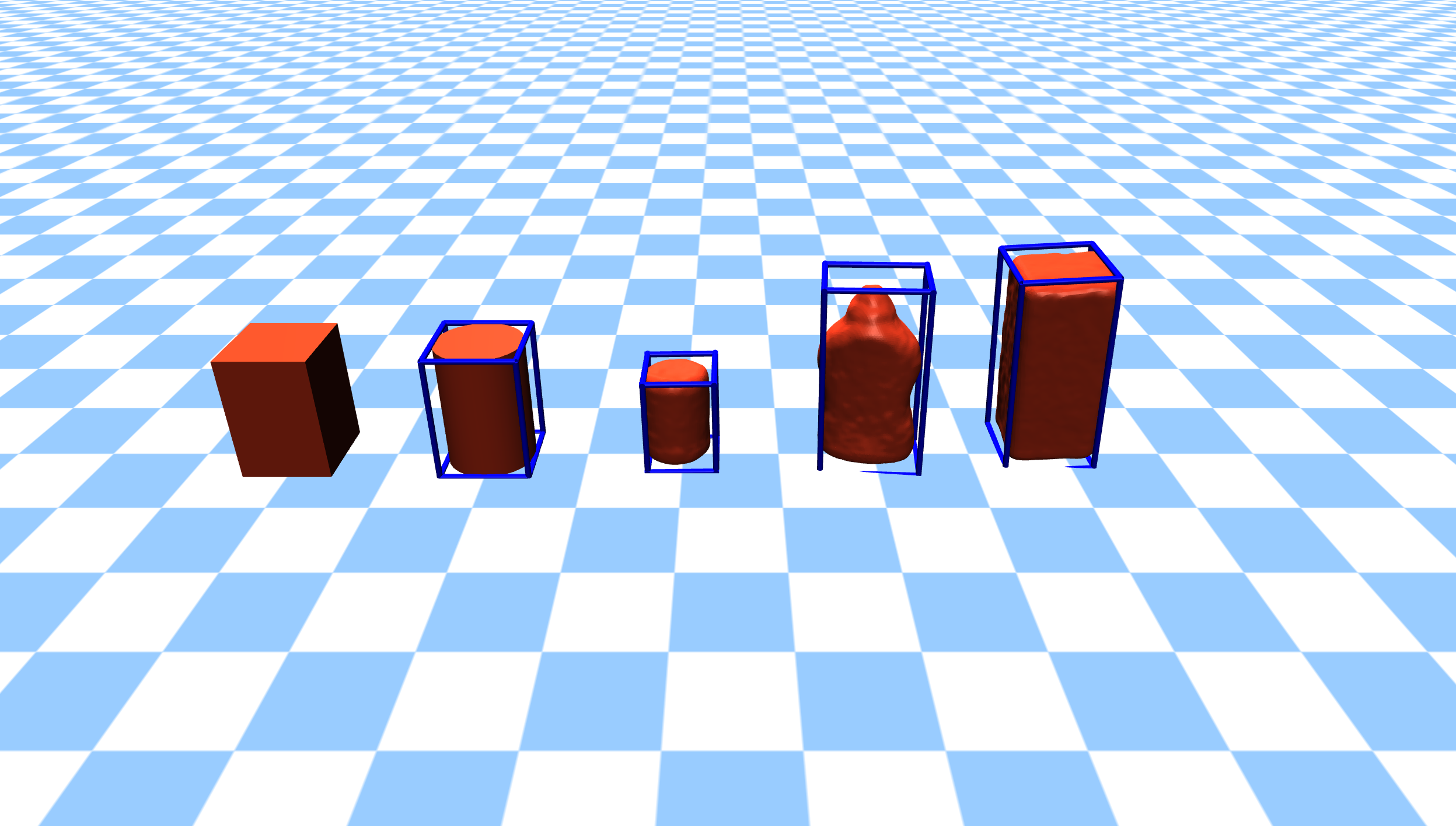}
	\caption{The training cube and testing objects with virtual bounding boxes.}
	\label{fig:new_objects}
	\vspace{-6mm}
\end{figure}

In this paper, for policy training, we use a cube as the grasping target. We utilize geometry key points, which can be seen as a sparse point cloud, to convey the object surface information. The key points consist of vertexes, centre of facets, centre of edges and the geometry centre, totaling 27 points. The geometry centre is estimated from other key points. For the testing objects displayed in \cref{fig:new_objects}, we utilize the geometry key points of their bounding boxes as the observation information. Acquisition of the bounding box and geometry key points can be achieved with Computer Vision methods \cite{2017_Han}. Considering the difficulty of obtaining complete point cloud in real experiment, we only utilize the \textbf{partial point cloud} of the object. We also introduce \textbf{sensor noises} into the positions of geometry key points. More details are in \cref{seq:robustness_visual}.

\begin{figure}[htbp]
\centering
\captionsetup{width=\linewidth}
\includegraphics[width=\linewidth]{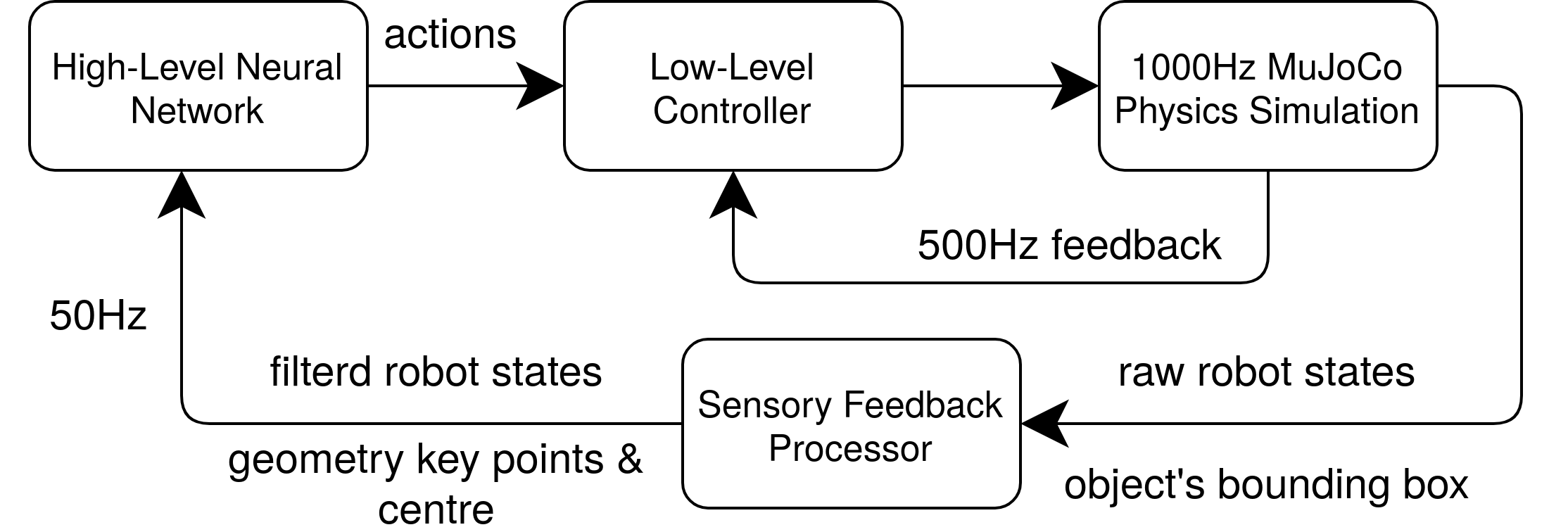}
\caption{Block diagram of the control framework.} 
\label{fig:control_framework}
\vspace{-5mm}
\end{figure}

\subsection{Control Framework} \label{seq:control_framework}
The control framework is designed in a hierarchical architecture, consisting of a high-level and a low-level loop, as shown in \cref{fig:control_framework}. The sensory feedback processor filters the raw robot states and extracts the geometry key points from the object's bounding box, estimating the object's geometry centre. The high-level controller is responsible for producing actions based on the processed environment state, with a frequency of 50Hz. Given the actions output from high-level controller, the 500Hz low-level proportional-derivative (PD) controller is responsible for computing target hand pose and velocity, as well as the finger joint torques, and feed them into the physics simulation.

\subsection{State and Action Space} \label{seq:state_action_space}
We use the term hand as the execution entity for reaching motion and fingers as the entity for grasping. To learn the synergy between hand and fingers, the agent needs to have full awareness of hand and fingers states. Therefore it takes the state $\mathcal{S}=\{\bm{X_r}, \theta, \bm{q}, \bm{d}, \bm{F}\}$ as input where $\bm{X_r}$ refers to the three dimensional object position which is estimated from object's bounding box; $\theta$ refers to the hand rotation angle around $Z$ axis; $\bm{q}$ refers to the three dimensional finger joint angles; $\bm{d}$ refers to the distances between each fingertip and the nearest object key point, which we find helpful when the hand is close to the object; $\bm{F}$ refers to the contact force measurements. Leveraging contact forces feedback in learning can improve the performance of learned grasp \cite{2019_ICRA_mbkrb}. The force sensors are attached on the inner hand as shown in \cref{fig:simulation_setup}, which will return the magnitude of the contact force.

The grasp detection problem concerns choosing the proper grasp pose and contact points based on the object's shape and that is beyond the scope of this paper. Therefore, to focus on the reactive control of reaching, grasping and re-grasping, we limit the DoF of the end-effector so that it can only approach and grasp the object laterally. We set the palm to always facing the lateral way, and constrain the translational motion of end-effector to the $XY$ two dimensional plane at a certain height. Only the rotation around $Z$ axis is allowed. Hence, the action $a$ consists of the hand translational velocities, rotational velocity around $Z$ axis, and the finger torques.

Considering the implementability of the policy in the real world, the working space is bounded within the reach of the robot arm. The maximum end-effector velocity is bounded within 1m/s and the finger torque is bounded within 2N$\cdot$m.

\begin{figure}[t]
	\centering
	\begin{subfigure}{0.45\linewidth}
		\centering
		\includegraphics[trim=220 100 50 100,clip,width=0.9\linewidth]{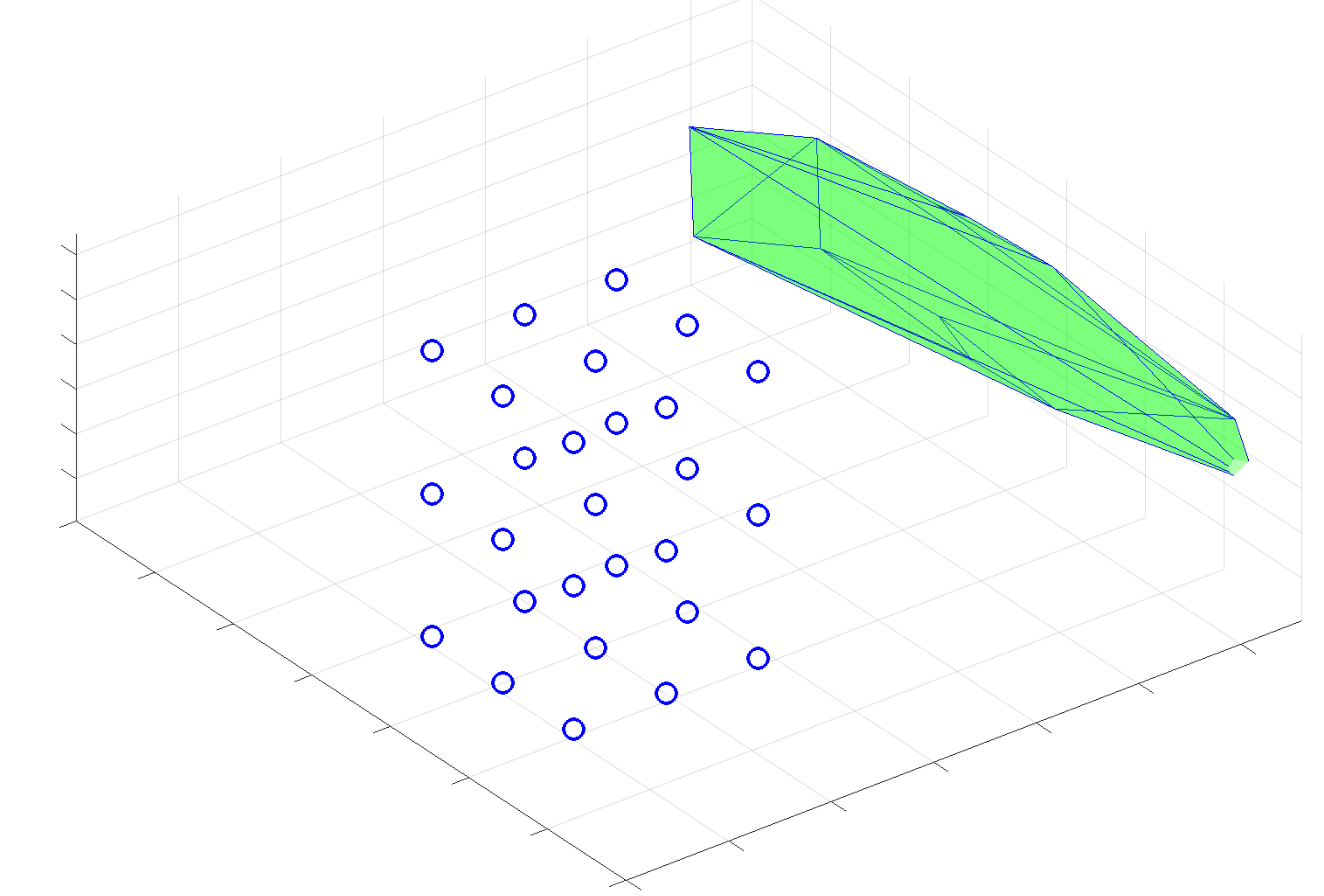}
		\caption{Reaching. $r_{\text{topology}}=0$}
		\label{fig:topology_reward1}
	\end{subfigure}
	\begin{subfigure}{0.45\linewidth}
		\centering
		\includegraphics[trim=200 100 300 260,clip,width=0.9\linewidth]{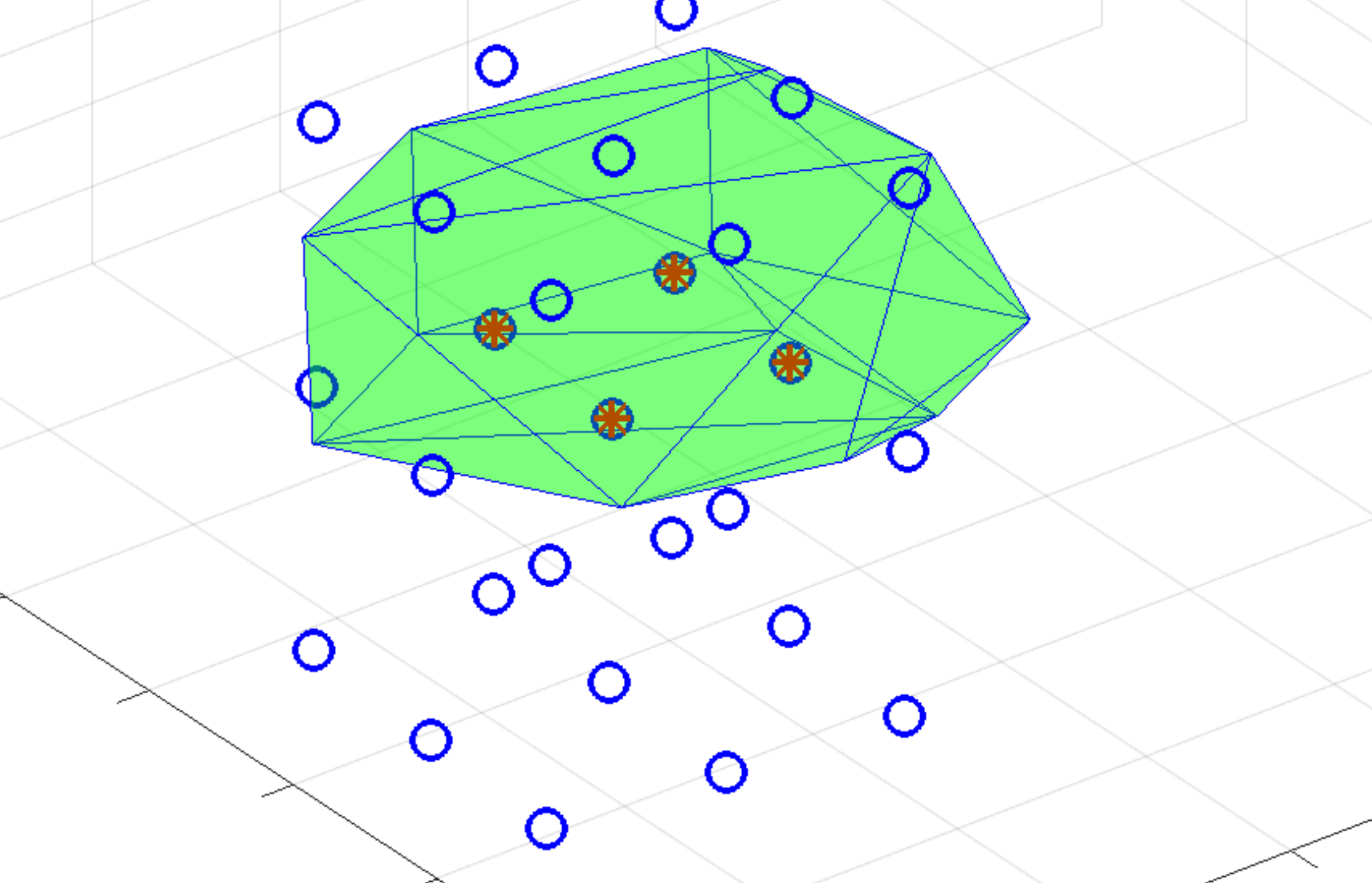}
		\caption{Grasping. $r_{\text{topology}}=\frac{4}{27}$}
		\label{fig:topology_reward2}
	\end{subfigure}
	\caption{Visualization of computing the topology reward $r_{\text{topology}}$, in reaching and grasping. Blue points: object geometric key points. Green area: the hand convex hull. In \cref{fig:topology_reward2} four object key points are inside the convex hull.} 
	\label{fig:topology_reward}
	\vspace{-5mm}
\end{figure}

\subsection{Reward Design} \label{seq:reward_design}
Reward design is one of the most important aspects in learning a good policy. With a poorly designed reward function, the learning may not converge to the desired policy and may lead to bad performance and safety issues \cite{2016_Amodei}. Reward design is a way to guide the policy search with the researchers' prior knowledge. From experience of how we reach and grasp things like mugs from a table, and observations on how infants learns to reach and grasp, we propose the following reward function, which is the linear combination of multiple positive reward terms and negative penalty terms with corresponding weights $\omega_i, i\in \{1,2,...,,6\} $:
\begin{equation} \label{eq:reward_function}
\begin{split}
  r &= \omega_1r_{\text{distTips}} + \omega_2r_{\text{vector}} + \omega_3r_{\text{contact}} \\
    &+ \omega_4r_{\text{topology}} + \omega_5p_{\text{collision}} + \omega_6p_{\text{objVel}}
\end{split},
\end{equation}
where $(\omega_1,\omega_2,\omega_3,\omega_4,\omega_5,\omega_6)=(1,1,2,10,-1,-2)$. The different terms are computed by the following equations.

The term $r_{\text{distTips}}$ rewards the distance between the hand and the object, and guides the agent learning to approach the object, where $X_i$ refers to the positions of hand key points - three fingertips and centre of the palm; $N_\text{p}$ refers to number of the object key points and $Y_j$ refers to their positions:
\begin{equation} \label{eq:reward_distTips}
r_{\text{distTips}} =  \text{exp}\left(-\sum_{i=1}^4\left(\min_{j\in\{1,..N_{\text{p}}\}}\norm{\bm{X}_i-\bm{Y}_j}\right)\right).
\end{equation}

In $r_{\text{vector}}$, the vector $\Vec{U_i}$ refers to the unit vector pointing from the hand key points to the estimated object geometry centre; $\Vec{N_i}$ refers to the normal vector of hand key points. The dot product will lead the hand to learn to face towards the object and grasp it in a proper direction:
\begin{equation}\label{eq:reward_vector}
r_{\text{vector}} = \frac{1}{4}\sum_{i=1}^4\left(\left(\Vec{U_i}\cdot\Vec{N_i}\right)\right).
\end{equation}

In the topology reward $r_{\text{topology}}$, $N_\text{p}$ refers to the total number of object key points observed; $n_\text{in}$ refers to the number of points which are inside the three dimensional convex hull formed by the hand and fingers. The convex hull is formed by multiple points including fingertips, finger joints and four corners of the square palm. \cref{fig:topology_reward} shows the convex hull formed by the hand and fingers in reaching and grasping motions:
\begin{equation} \label{eq:reward_topology}
r_{\text{topology}} = \frac{1}{N_{\text{p}}}n_\text{in}.
\end{equation}

In addition to $r_\text{topology}$, a contact term $r_\text{contact}$ is added to encourage power (enveloping) grasping, which is more stable than precision (fingertip) grasp:
\begin{equation} \label{eq:reward_contact}
r_{\text{contact}} = n_{\text{con}},
\end{equation}
where $n_\text{con}$ refers to the number of contact points between object and hand. This term encourages more contact points of the hand with the object during the grasp, under the assumption that with more contact points, the more stable the grasp is. Note that only the contact points in the inner part of hand are counted in $n_{\text{ncon}}$. 

If the hand contacts the object with the outer side of fingers, the contact points are counted in $n_\text{c}$, which is regarded as the penalty term:
\begin{equation}
p_{\text{collision}} = n_{\text{c}}.
\end{equation}

A penalty $p_{\text{objVel}}$ on the object translational velocity is added to prevent the hand from pushing the object away and encourage a gentle grasping behavior:
\begin{equation}
p_{\text{objVel}} = \norm{\bm{V_{\text{obj}}}}.
\end{equation}

\begin{figure}[t]
	\centering
	\begin{subfigure}{0.45\linewidth}
		\centering
		\captionsetup{width=0.9\linewidth}
		\includegraphics[height=25mm]{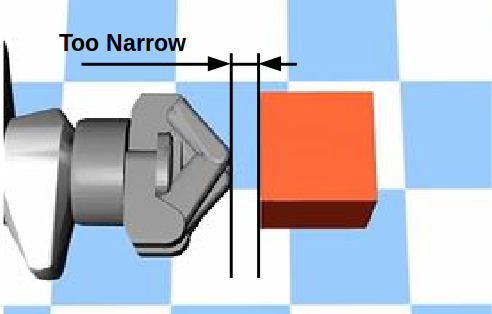}
		\caption{Potential collision.}
	\end{subfigure}
	\begin{subfigure}{0.45\linewidth}
		\centering
		\captionsetup{width=0.9\linewidth}
		\includegraphics[height=25mm]{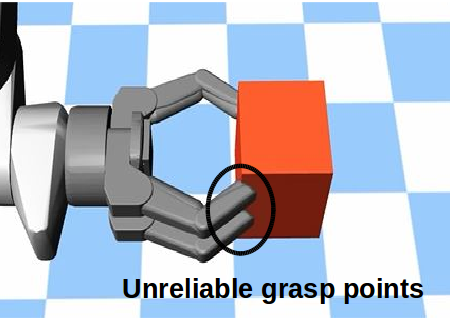}
		\caption{Unreliable grasp points.}
	\end{subfigure}
	\caption{Two challenging initial states during training.} 
	\label{fig:initial_states}
	\vspace{-4mm}
\end{figure}

\subsection{Learning Episode}
A single learning path lasts for 400 time steps in simulation. An episode starts with randomizing the object position, where the cube is being set at a random place on the ground within the operating range. In order to enable the hand to obtain the ability of grasping moving targets, we added random disturbances on the object. The candidate disturbances consist of forces applied on the centre of mass of object in four lateral directions. There is one disturbance duration in each episode, happening at any time in the first half of it, lasting for 0.3 seconds, so the object will slide towards the force direction if not being grasped already.

With randomly added disturbances, the agent can gather enough trials to learn to track and grasp moving target, but it usually fails to achieve the re-grasp motion if the object slips away during the grasp. Randomly distributed disturbance does not provide sufficient data points for the agent to learn the re-grasp. The re-grasp requires synergy motion of fingers and hand. To avoid collision between outer side of fingers and the object, the hand sometimes needs to move backward to make sure there is enough space for opening the fingers. Thus, apart from the normal training episode, we designed two initial states with special finger joints and object position to train the re-grasp policy, as shown in \cref{fig:initial_states}.

\section{Evaluation and Analysis} \label{seq:evaluation}
In this section, we present the results of the learned policy in simulation, and evaluate the capabilities of reaching, grasping and re-grasping with different metrics and tasks. Moreover, we discuss the necessity and effect of each reward term and initial training states. The evaluation tasks contain: (1) static grasp with random object position; (2) dynamic grasp where a force with random direction would be applied on the object for a certain duration; (3) re-grasp starting from initial configurations shown in \cref{fig:initial_states} and (4) dynamic re-grasp where the object would be moved out of the hand during first grasp attempt.

\begin{table}[t]
	\centering
	\caption{The evaluation of the policy in different tasks.}
	\label{tab:grasp_quality}	 
	\begin{tabular}{p{3cm} | >{\centering}p{0.5cm} >{\centering}p{0.6cm} >{\centering}p{0.6cm} >{\centering}p{1.2cm}}
		\hline
		& Lift & $\text{Shake}_{12}$ & $\text{Shake}_{15}$ & Recover$[s]$ \tabularnewline
		\hline
		Static Target Grasp & 97\% & 90\% & 69\% & $\varnothing$ \tabularnewline
		Dynamic Grasp($5N, 0.3s$) & 98\% & 88\% & 74\% & $\varnothing$ \tabularnewline
		Dynamic Grasp($8N, 0.3s$) & 78\% & 78\% & 60\% & $\varnothing$ \tabularnewline
		Close Fingers Re-grasp& 100\% & 92\% & 56\% & 0.92 \tabularnewline
		Shallow Grasp Re-grasp& 100\% & 100\% & 91\% & 0.69\tabularnewline
		Dynamic Re-grasp& 83\% & 79\% & 56\% & 1.48 \tabularnewline
		\hline 
	\end{tabular}
	\vspace{-3mm}
\end{table}


\begin{figure*}[htbp]
	\centering	
	\begin{subfigure}{0.45\linewidth}
		\centering
		\captionsetup{width=0.8\linewidth}
		\includegraphics[trim=0 250 0 250,clip,width=0.9\linewidth]{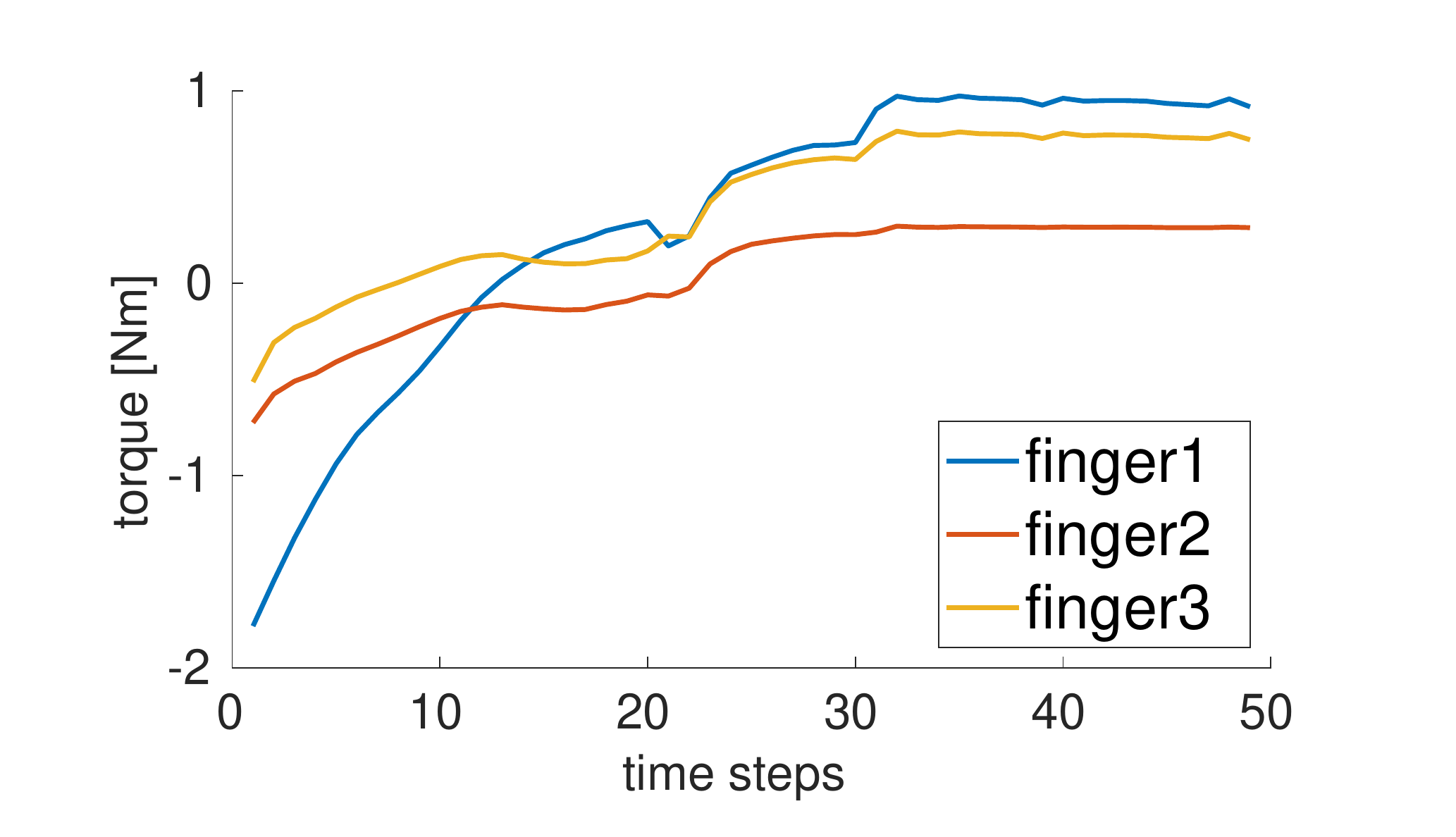}
		\caption{Torque command of each finger.}
		\label{fig:finger_action}
	\end{subfigure}
	\begin{subfigure}{0.45\linewidth}
		\centering
		\captionsetup{width=0.8\linewidth}
		\includegraphics[trim=0 250 0 250,clip,width=0.9\linewidth]{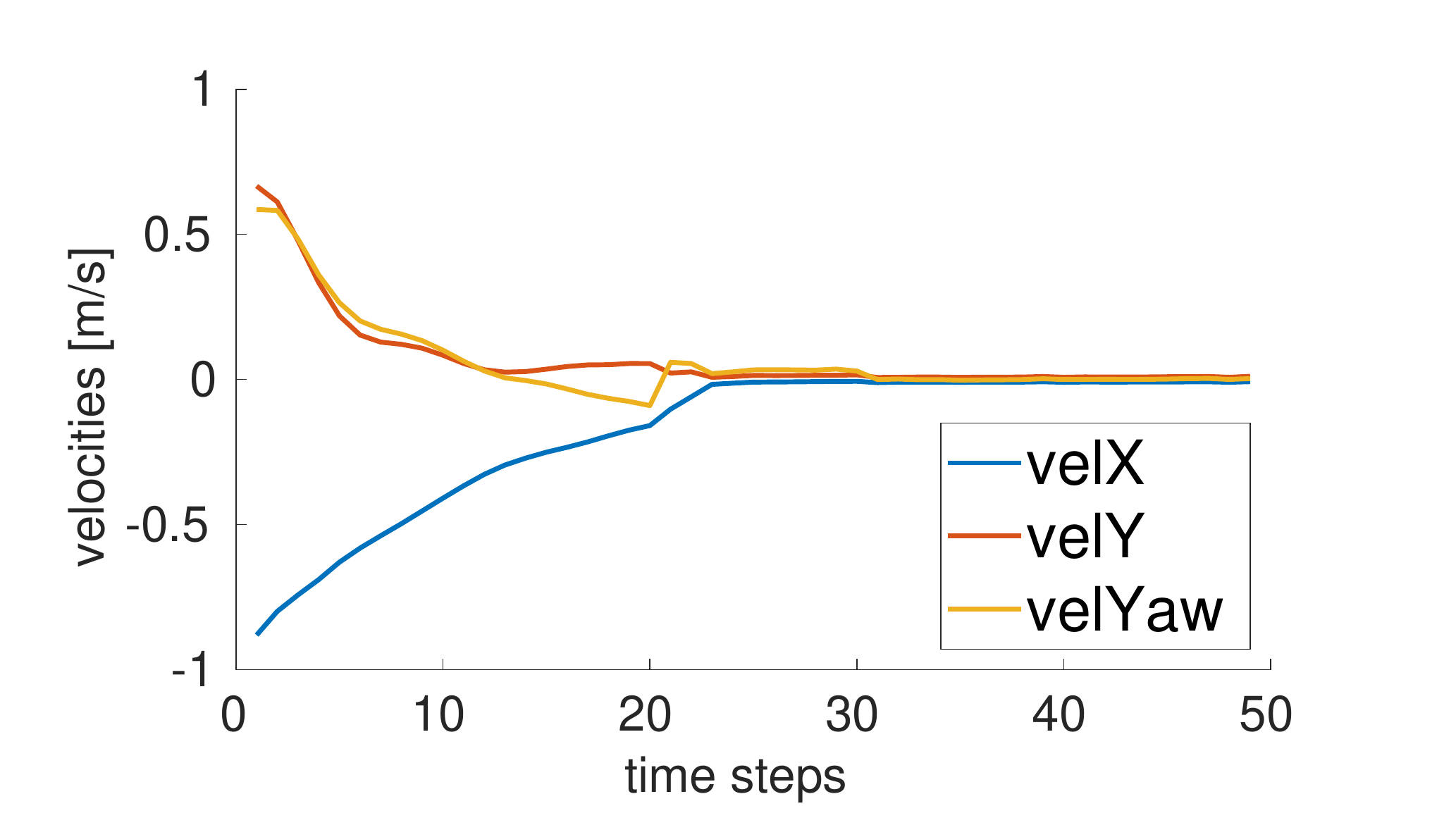}
		\caption{Translational velocities of the hand.}
		\label{fig:hand_action}
	\end{subfigure}
	\caption{The output actions of learned policy during one canonical trial of grasping a static object placed at a random position.} 
	\label{fig:action_static_object}
	\vspace{-2mm}
\end{figure*}

\subsection{Evaluation Metrics}
We evaluate the performance by two metrics: the success rate in lift-and-shake tests, and recovery time specially used for re-grasp motion. 

In the lifting test, the robot first lifts up the hand, and if the object can be held for ten seconds, we regard the test as a success. In the shaking test, the robot first lifts the hand and then disturbance forces with random directions are applied on the object. If the grasp lasts for ten seconds, this trail is marked as a success. Here we set the force to $12N$ and $15N$. We record the time that the policy spends to recover from the failure and accomplish a successful re-grasp attempt. The recovery time is the average time required for the agent to achieve a robust grasp which passes the lift-and-shake tests.

In \cref{tab:grasp_quality} we present the evaluation of different tasks with different metrics. The results indicate that our approach can generate a robust control policy which can react to changes rapidly and execute re-grasp in case of failures, even under difficult configurations as shown in \cref{fig:initial_states}. The achieved grasps are stable for lifting in most cases and can resist external disturbances to some extent.

\subsection{Static Target}\label{seq:static_target}
In this test, the object is placed at a random position, and the task is to reach and grasp it. The first row of \cref{tab:grasp_quality} indicates that the learned policy is able to achieve a stable grasp for a static target.

\cref{fig:action_static_object} demonstrates the actions in a typical static grasp task. When the hand is relatively far from the object, the finger torques are negative, which means the fingers are extending and open, enlarging the grasping area. The moving velocity of hand reaches the maximum at the beginning and converge to zero at the end, indicating that the agent learns to slow down when the hand approaches the object. This behaviour coincides with how humans would grasp an object.

\subsection{Moving Target}\label{seq:moving_target}
In reinforcement learning, at every control step, the agent takes the current observation of the environment as input and outputs the corresponding actions. In this paper the agent has no perception of the object's motion status such as velocity or acceleration. Therefore it is unable to predict the object's future position. Without knowing the object velocity and acceleration, the agent will not be able to learn an optimal policy of grasping a moving object. However, due to the randomization of object position and the disturbances applied on the object during the training, combined with a high enough control frequency. The learned sub-optimal policy has decent tracking capability and is able to dynamically re-adjust to grasp a moving object as long as the object's velocity is within the agent's operational velocity.

\cref{fig:moving_grasp} demonstrates the motion of grasping a moving object. In this test setting, the object locates at a random position. A force pointing to a random direction within the $X-Y$ plane is applied on the object's centre for 0.3 second. So the object will start moving at the beginning of the trial. According to \cref{tab:grasp_quality}, the learned policy is able to adjust online based on the change of object position.

\begin{figure*}[t]
	\centering\offinterlineskip
	\begin{subfigure}{\linewidth}
		\centering
		\captionsetup{width=0.95\linewidth}
		\includegraphics[trim=10 0 0 0,clip,width=0.95\linewidth]{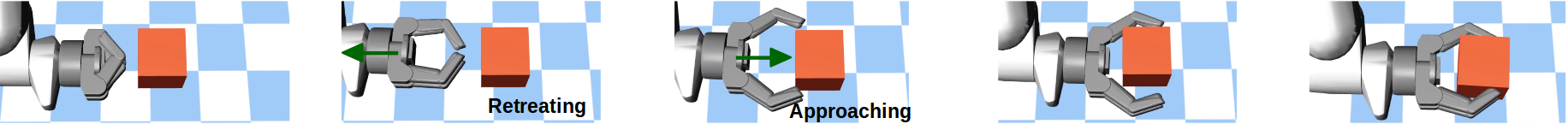}
		\caption{Collision-free re-grasping from the initial configuration where the object is close to the fingers.}
		\label{fig:closed_regrasp}
		\vspace{1.5mm}
	\end{subfigure}
	\begin{subfigure}{\linewidth}
		\centering
		\captionsetup{width=0.95\linewidth}
		\includegraphics[width=0.95\linewidth,height=13.4mm]{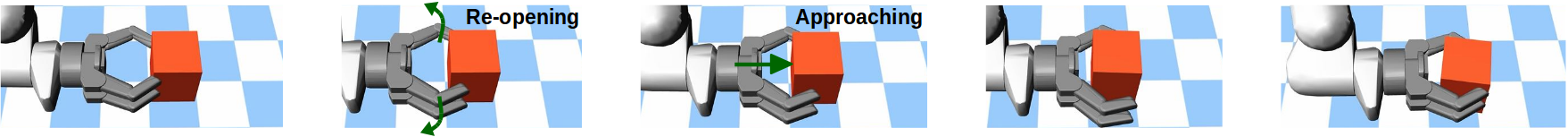}
		\caption{Collision-free re-grasping with accurate finger-hand coordination from the initial configuration of Unreliable grasp.}
		\label{fig:shallow_regrasp}
		\vspace{1.5mm}
	\end{subfigure}
	\begin{subfigure}{\linewidth}
		\centering
		\captionsetup{width=0.95\linewidth}
		\includegraphics[width=0.95\linewidth,height=13.4mm]{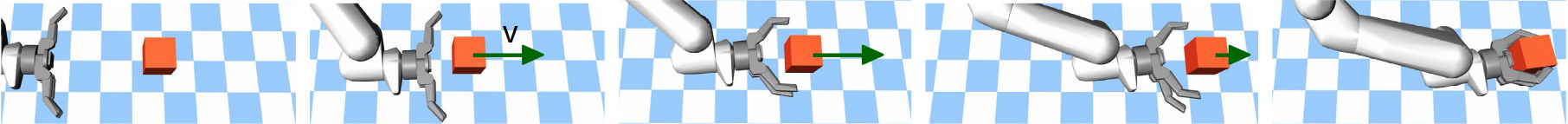}
		\caption{Collision-free re-grasping of a dynamically moving object.}
		\label{fig:dynamic_regrasp}
	\end{subfigure}
	\caption{The snapshots of re-grasping motions generated from the learned policy.}
	\label{fig:regrasp_snapshots}
\end{figure*}

\subsection{Re-grasp Test}\label{seq:regrasp_test}
In this test setting, in order to evaluate the ability of re-opening the fingers and applying another grasp attempt, the object will be moved out of the hand at the timing when the distance between fingertips and the object is below the threshold. There are two potential consequences. First, the object moves out of the cage formed by fingers and palm completely. Then the fingers will stop closing and re-open. If the object blocks the fingers from opening, the hand needs to retract backwards to create enough space before re-grasping. In the second situation, the object is caught by the fingertips while moving. This will cause a Unreliable shallow grasp, so the agent learns to release and re-grasp the object. 

\cref{fig:closed_regrasp}, \cref{fig:shallow_regrasp} display the learned re-grasping motions from two aforementioned challenging configurations. In \cref{fig:dynamic_regrasp}, the object suddenly moves away to right since the second snapshot, and as a responsive coordination, the fingers re-open and the hand chases the object until executing another grasping attempt in close proximity. The agent learns an effective manner to return to a proper pre-grasp posture without any collision and redundant motions. According to the results listed in \cref{tab:grasp_quality}, the learned policy is able to recover from failures within 2 seconds, which is far more efficient than resetting the whole control pipeline, and achieve a stable grasp which can pass the lift-and-shake tests with high success rates.

\begin{figure}[t]
	\centering
	\captionsetup{width=\linewidth}
	\includegraphics[width=1.0\linewidth]{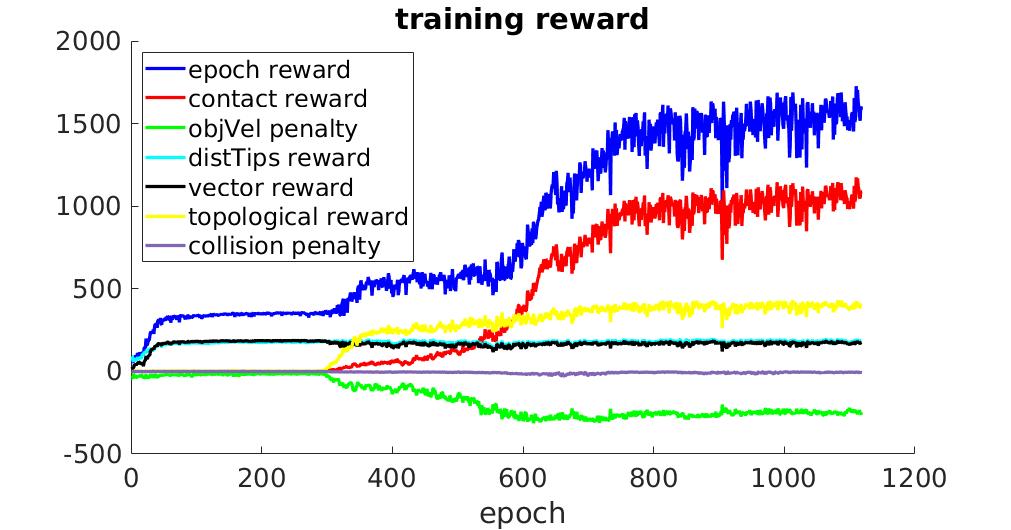}
	\caption{The learning curves in a typical learning process. The blue curve is the sum of all reward terms.}
	\label{fig:learning_curve}
\end{figure}

\subsection{Ablation Study}\label{seq:ablation_study}
In order to show the necessity and effectiveness of each reward term, we remove them from the reward function, and compare the learned policies with and without them. \cref{fig:learning_curve} demonstrates the learning curves of each reward term, from which we can tell that every term contributes in the policy learning. \cref{tab:ablation_study} demonstrates the capabilities of different policies trained with incomplete reward function.

The $r_{\text{distTips}}$ and $r_{\text{vector}}$ affect the learning first and guide the agent to approach the target. Without $r_{\text{distTips}}$, the hand cannot learn to reach the object. $r_{\text{vector}}$ guides the fingers to keep open when approaching the target. After removing it, the agent fails to learn the proper grasping motion.

The $r_{\text{topology}}$ and $r_{\text{contact}}$ come to influence afterwards and mainly guide the learning of grasping motion. $r_{\text{topology}}$ plays an important role in the early stage of grasping learning after the hand learns to approach the object. It guides the fingers to wrap around the object, and also compensates the penalty on object velocity caused by the random actions from policy search. After removing $r_{\text{topology}}$, the hand learns to stay at a certain distance from the object with fingers open, in order to avoid the penalty on object velocity. $r_{\text{contact}}$ encourages the palm and fingers to contact the object and leads to a firm grasp. After removing $r_{\text{contact}}$, the hand learns to stay at a closer distance from the object and the fingers will form a cage around it, instead of contacting and grasping it.

The penalty on object velocity $p_{\text{objVel}}$ is responsible for achieving gentle grasp motion and preventing the hand from moving the object. The agent can still learn the reaching and grasping without this reward term, but the hand will move the object randomly after a successful grasp. The collision penalty $p_{\text{collision}}$ and the two special initial states in \cref{fig:initial_states} are essential for learning the collision-free re-grasping. With $p_{\text{collision}}$ and initial states, the hand will move backwards to create enough space for the fingers to open. After removing them, the hand will not move backwards and the outer part of fingers will collide with the object while opening.

\begin{table}[t]
	\centering
	\caption{Capability tests of policies learned with incomplete reward functions.}
	\label{tab:ablation_study}	  
	\def\arraystretch{1.3}
	\begin{tabular}{l | c c c c}
		\hline
		& Reach & Grasp & Re-grasp & Lift \tabularnewline
		\hline
		No $r_{\text{distTips}}$ & \xmark & \xmark & \xmark &\xmark\tabularnewline
		No $r_{\text{vector}}$ & \cmark & \xmark & \xmark &\xmark\tabularnewline
		No $r_{\text{topology}}$ & \cmark & \xmark & \xmark &\xmark\tabularnewline
		No $r_{\text{contact}}$ & \cmark & \xmark & \xmark &\xmark\tabularnewline
		No $p_{\text{objVel}}$ & \cmark & \cmark & \cmark &\cmark\tabularnewline
		No $p_{\text{collision}}$ & \cmark & \cmark & \xmark &\cmark\tabularnewline
		No special initial states & \cmark & \cmark & \xmark &\cmark\tabularnewline
		\hline 
	\end{tabular}
	\vspace{-3mm}
\end{table}

\subsection{Evaluation on unseen objects}

In order to evaluate the policy's robustness and generalization ability, we repeat the static grasping task with four different unseen objects in \cref{fig:new_objects}, including a cylinder, a can, a mustard bottle and a wood block. The meshes of the objects are imported from the YCB database \cite{2015_YCB}. Since the policy takes the geometry key points on processed bounding box of the target object, as well as the in-hand contact forces, as input, it shows promising capability of grasping columnar objects with irregular shapes and various sizes.

\begin{table*}[t]
	\centering
	\caption{Peak finger torques and hand velocities for different scenarios.}
	\label{tab:peak_toruqe_vel}	  
	\def\arraystretch{1.3}
	\begin{tabular}{p{3.0cm} >{\centering}p{1.8cm} >{\centering}p{1.8cm} >{\centering}p{1.8cm} | >{\centering}p{1.8cm} >{\centering}p{1.8cm} >{\centering}p{1.8cm}}
		\hline
		& \multicolumn{3}{c}{Peak Finger Joint Torque [$N\cdot m$]} & \multicolumn{3}{c}{Peak Hand Velocity}\tabularnewline
		\cline{2-7}
		& Finger 1 Base & Finger 2 Base & Finger 3 Base & X [$m/s$] & Y [$m/s$] & Yaw [$\text{rad}/s$]\tabularnewline
		\hline
		Limit & 2 & 2 & 2 & 1 &1 & 1 \tabularnewline
		\hline
		Static Grasp & 1.81 & 1.01 & 1.52 & 0.86 & 0.81 & 0.72\tabularnewline
		Dynamic Grasp($5N, 0.3s$) & 1.59 & 0.77 & 1.08 & 0.75 & \bf{1} & 0.95\tabularnewline
		Dynamic Grasp($8N, 0.3s$) & 1.69 & 0.78 & 0.96 & 0.78 & 0.96 & 0.81\tabularnewline
		Close Fingers Re-grasp& 1.49 & 1.91 & 1.71 & 0.75 & \bf{1} & 0.91 \tabularnewline
		Shallow Grasp Re-grasp& 1.12 & 0.91 & 0.99 & 0.81 & 0.62 & 0.54 \tabularnewline
		\hline 
	\end{tabular}
\end{table*}

\begin{figure}[t]
	\centering
	\begin{subfigure}{0.45\linewidth}
		\centering
		\captionsetup{width=0.8\linewidth}
		\includegraphics[trim=200 100 150 200,clip,width=0.7\linewidth]{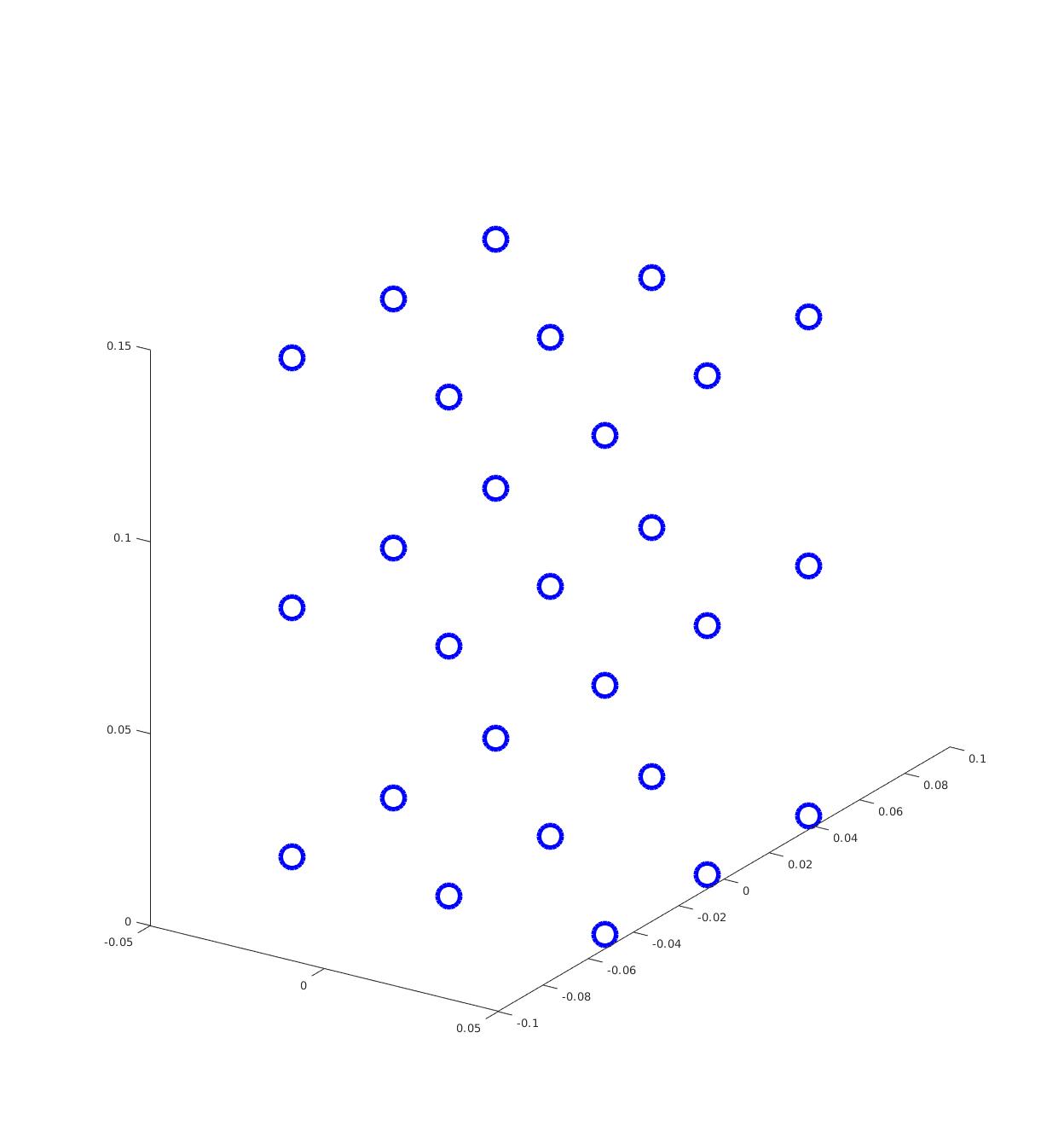}		
		\caption{}
		\label{fig:complete_point_cloud}
	\end{subfigure}
	\begin{subfigure}{0.45\linewidth}
		\centering
		\captionsetup{width=0.8\linewidth}
		\includegraphics[trim=200 100 150 200,clip,width=0.7\linewidth]{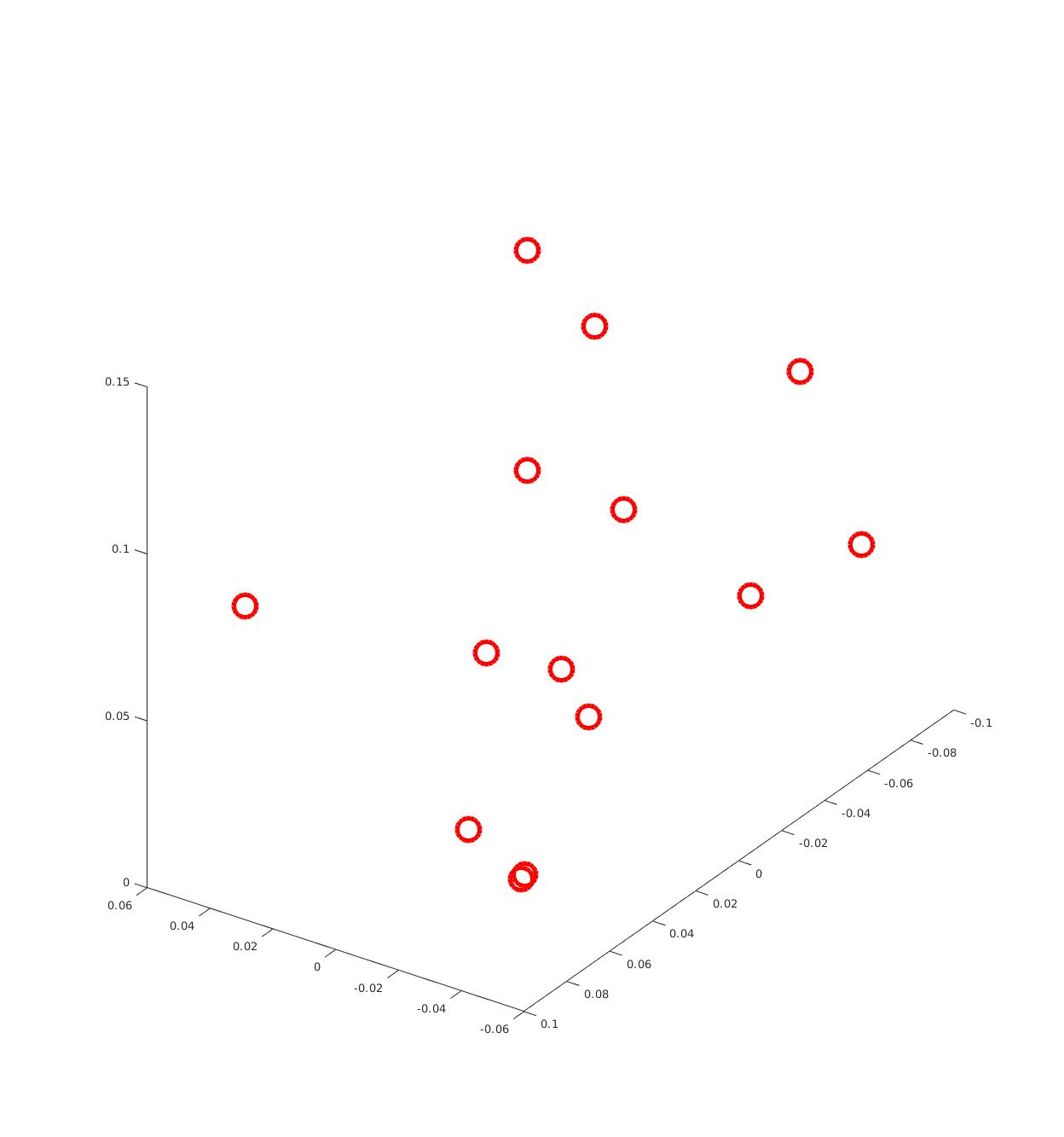}		
		\caption{}
		\label{fig:noisy_pc0}
	\end{subfigure}
	\begin{subfigure}{0.45\linewidth}
		\centering
		\captionsetup{width=0.8\linewidth}
		\includegraphics[trim=150 50 100 150,clip,width=0.7\linewidth]{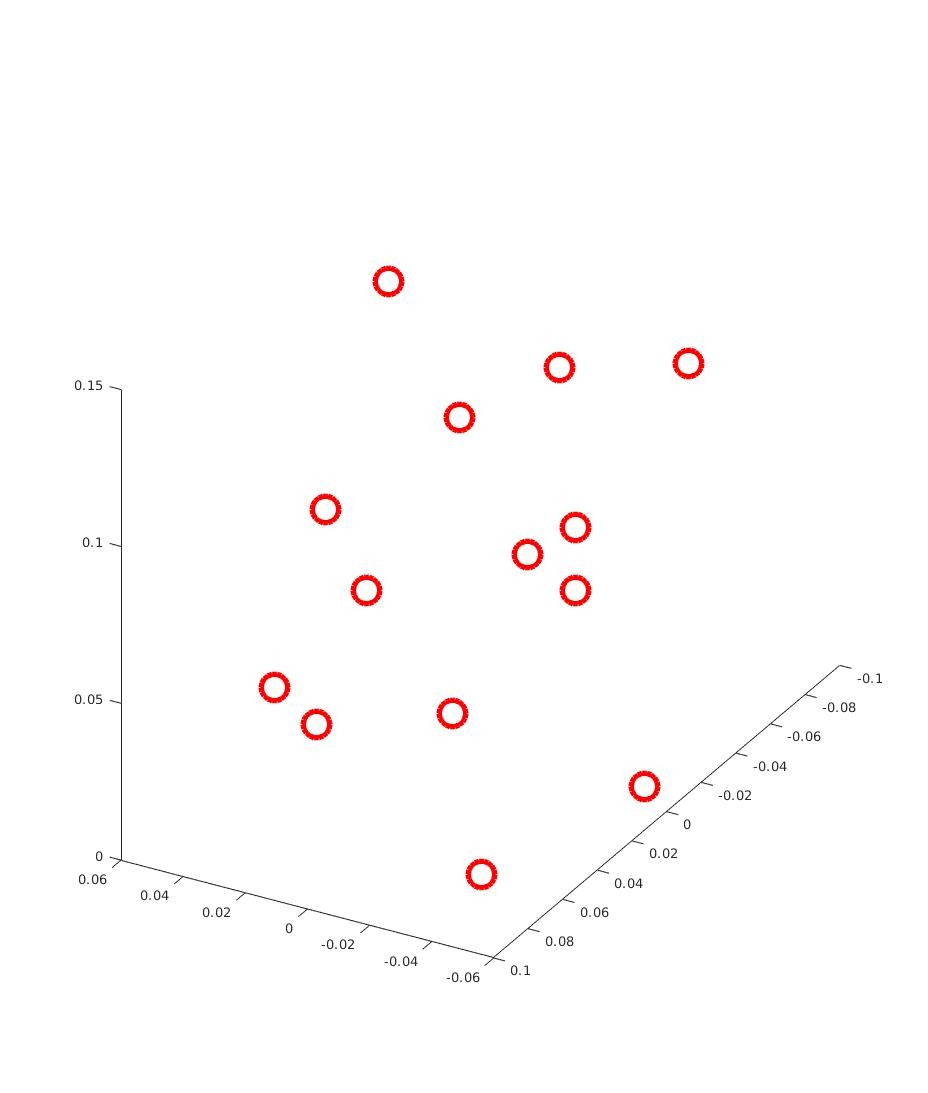}		
		\caption{}
		\label{fig:noisy_pc1}
	\end{subfigure}
	\begin{subfigure}{0.45\linewidth}
		\centering
		\captionsetup{width=0.8\linewidth}
		\includegraphics[trim=150 50 100 150,clip,width=0.7\linewidth]{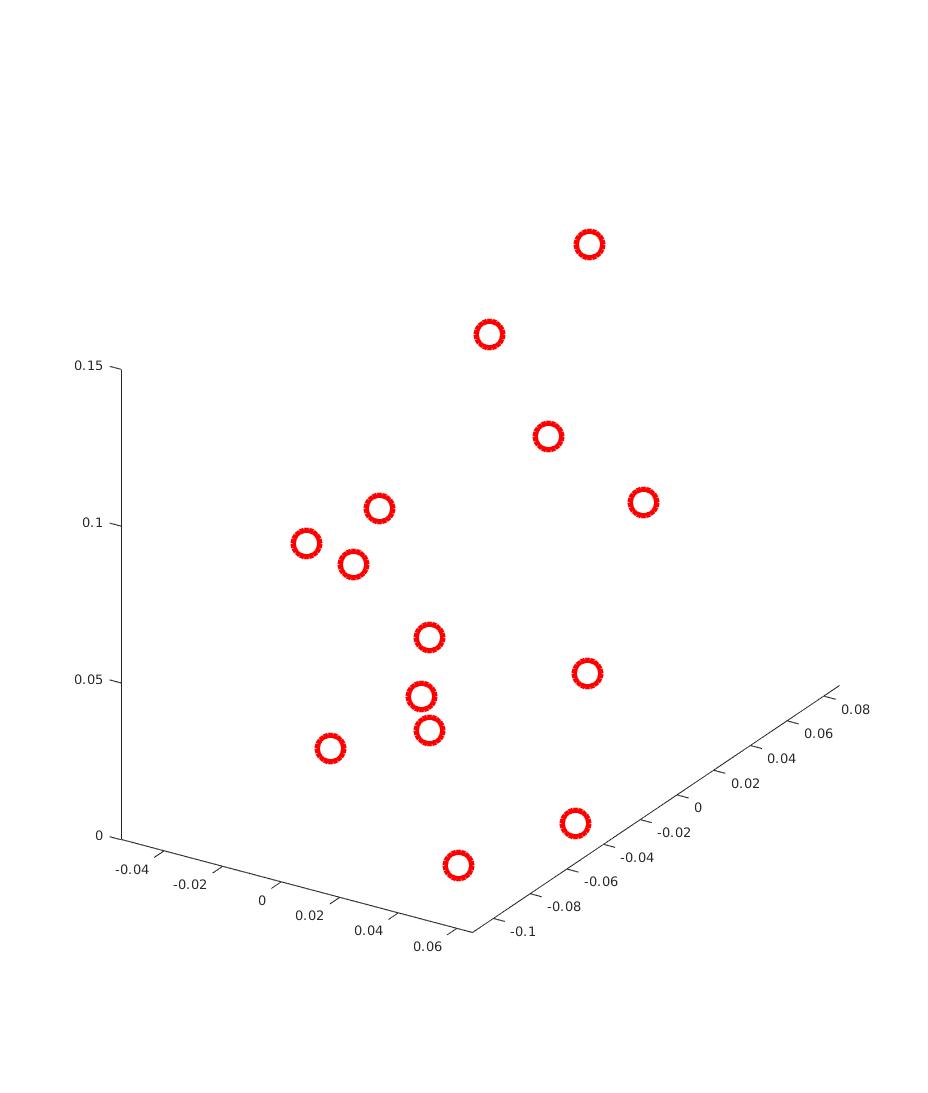}		
		\caption{}
		\label{fig:noisy_pc2}
	\end{subfigure}
	\caption{Comparison of ideal and realistic noisy partial observations of object's key points: (a) complete, noisy-free point cloud; (b-d) randomized noisy partial observations used in simulation for training and testing.}
	\label{fig:noisy_partial_pc}
	\vspace{-4mm}
\end{figure}

\subsection{Robustness on Partial Visual Observation}\label{seq:robustness_visual}

In real world experiments, it is hard to obtain the noisy-free complete point cloud of the target. Therefore, considering the sim-to-real transfer, during the policy training, we only utilize half of the geometry key points to encode object's surface information and estimate the object position. We also introduce sensory noises to the key points positions, which are sampled from a normal distribution with $\mu=0, \sigma=0.02m$. \cref{fig:noisy_partial_pc} demonstrates the comparison of the complete point cloud of the cube and an example of noisy partial point cloud applied in the policy training. The results suggest that the learned policy is robust to partial noisy visual input.

\subsection{Feasibility of Hardware Experiment}

Although the training and evaluation tests are conducted in simulation, we take the realism of generated policy into consideration by discouraging aggressive motions through penalties and restrict the hand velocities and finger torques. We also purposely introduce sensor noises in the state observation and only use partial key points of the object for the feedback. Therefore, these realistic setting in training paves a way for the sim-to-real transfer. \cref{tab:peak_toruqe_vel} displays the average peak finger joint torques and hand velocities in different evaluation tasks. From the table we can ensure that the motions generated from learned policy are within the constraints and have enough safety margin. For implementation on different hardware platform, we only need to match the limits used in simulation with those of the real robots.

\section{Conclusion} \label{seq:conclusion}
In this paper, we used model-free reinforcement learning to acquire the combined reactive control policy of reaching and grasping. In simulation, we used a cube as the target for grasping, a three-fingered robot hand as the end-effector. The agent explores and optimizes the policy through trial-and-error, guided by a well-defined reward function. Apart from the initial learning state with random object configuration, we also incorporate two challenging initial states to train the re-grasping ability by inducing failed attempts. The training results showed that the learned agent can reach and grasp a static target, and also grasp a moving object and generate a collision-free re-grasp online after a failure. Through ablation experiments, we demonstrated how each reward term and the special initial states improve the capability and robustness of the learned policy.

As for future work, we will use a real robotic hand mounted on the Franka robot arm to transfer the learned policy into reality, which is also considered in our training phase as we applied the constraints on the hand motion restricted by the Franka arm. Further research will also be done on robot perception to employ more complicated visual feedback into the reinforcement learning framework.
\bibliography{DRG-wenbin}
\end{document}